\pdfoutput=1
\documentclass[10pt,twocolumn,letterpaper]{article}

\newif\ifarxiv
\arxivtrue

\ifarxiv
\usepackage[pagenumbers]{cvpr} %
\else
\usepackage{cvpr}              %
\fi

\def\MYTITLE{Event-based Continuous Color Video Decompression from Single Frames} %

\usepackage[dvipsnames]{xcolor}

\def\bx{\mathbf{x}}

\def\bu{\mathbf{u}}

\def\mB{\mathtt{B}}

\def\Real{\mathbb{R}}

\def\numPoints{N_p} %
\def\numBases{N_b} %
\def\numTimestamps{N_t} %
\def\f{\mathbf{f}} %
\def\cI{\mathcal{I}} %
\def\bq{\mathbf{q}} %
\def\featplane{\mathbf{g}} %

\usepackage{graphicx}
\usepackage{booktabs}

\usepackage{xcolor}
\definecolor{claude_red}{RGB}{255,0,0}

\definecolor{ken_green}{RGB}{0,255,0}

\usepackage{multicol}
\usepackage{dirtree}

\usepackage{pifont}
\usepackage{subcaption}
\usepackage{caption}
\usepackage{array}
\usepackage{graphicx}
\usepackage{makecell}
\usepackage{rotating}
\usepackage{adjustbox}

\definecolor{light-gray}{gray}{0.6}

\newcommand{\cmark}{\ding{51}}%
\newcommand{\xmark}{\ding{55}}%
\newcommand{\ours}[1]{ContinuityCam}
\newcommand{\ourdatafull}[1]{Event Extreme Decompression Dataset}
\newcommand{\ourdata}[1]{E2D2}
\newcommand{\bnum}[1]{\bfseries #1}

\newif\ifshowmainpaper %
\showmainpapertrue
\newif\ifshowsupplementary
\showsupplementarytrue

\definecolor{cvprblue}{rgb}{0.21,0.49,0.74}
\ifarxiv
\usepackage[breaklinks,colorlinks,allcolors=cvprblue,bookmarks=true,bookmarksnumbered=true]{hyperref}
\hypersetup{
  pdftitle={\MYTITLE},
  pdfsubject={Computer Vision, Deep Learning},
  pdfauthor={Ziyun Wang, Friedhelm Hamann, Kenneth Chaney, Wen Jiang, Guillermo Gallego, Kostas Daniilidis},
  pdfkeywords={Event Cameras, Video Synthesis}
}
\usepackage[absolute]{textpos}
\else
\usepackage[pagebackref,breaklinks,colorlinks,allcolors=cvprblue]{hyperref}
\fi

\def\paperID{22} %
\def\confName{CVPR}
\def\confYear{2025}

\title{\MYTITLE}

\author{Ziyun Wang$^{1}$, Friedhelm Hamann$^{2}$, Kenneth Chaney$^{1}$, Wen Jiang$^{1}$, \\
Guillermo Gallego$^{2}$, Kostas Daniilidis$^{1,3}$\\
$^{1}$University of Pennsylvania, USA.\\
$^{2}$TU Berlin, ECDF Berlin, SCIoI Cluster and RIG, Germany.
$^{3}$Archimedes, Athena RC
\\
}

\begin{document}

\ifarxiv
\definecolor{somegray}{gray}{0.5}
\newcommand{\darkgrayed}[1]{\textcolor{somegray}{#1}}
\begin{textblock}{10.5}(2.8, 0.7)  %
\begin{center}
\darkgrayed{This paper has been accepted for publication at the
IEEE/CVF Conference on Computer Vision and Pattern Recognition (CVPR) Workshops, Nashville (USA), 2025.
\copyright IEEE
}
\end{center}
\end{textblock}
\fi

\maketitle

\ifshowmainpaper
\maketitle
\begin{abstract}

We present\emph{~\ours{}}, a novel approach to generate a continuous video from a single static RGB image and an event camera stream. 
Conventional cameras struggle with high-speed motion capture due to bandwidth and dynamic range limitations. 
Event cameras are ideal sensors to solve this problem because they encode compressed change information at high temporal resolution. 
In this work, we tackle the problem of event-based continuous color video decompression, pairing single static color frames and event data to reconstruct temporally continuous videos.
Our approach combines continuous long-range motion modeling with a neural synthesis model, enabling frame prediction at arbitrary times within the events. 
Our method only requires an initial image, thus increasing the robustness to sudden motions, light changes, minimizing the prediction latency, and decreasing bandwidth usage. 
We also introduce a novel single-lens beamsplitter setup that acquires aligned images and events, and a novel and challenging \ourdatafull{} (\ourdata{}) that tests the method in various lighting and motion profiles. 
We thoroughly evaluate our method by benchmarking color frame reconstruction,
outperforming the baseline methods by 3.61 dB in PSNR and by 33\% decrease in LPIPS,
as well as showing superior results on two downstream tasks. 
Please see our project website for details: \url{https://ziyunclaudewang.github.io/continuitycam/}.

\end{abstract}

\section{Introduction}
\label{sec:intro}
\begin{figure}[tb]
    \centering
    \includegraphics[clip, trim={4.8cm, 12.9cm, 11cm, 4cm}, width=\linewidth]{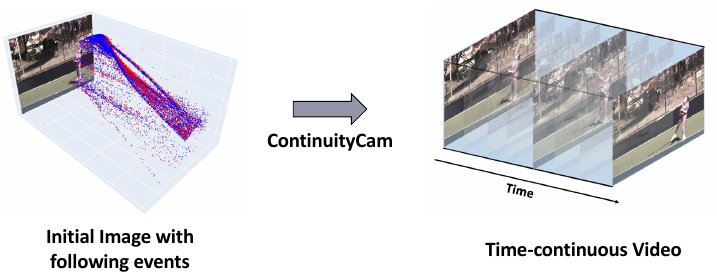}
    \caption{\label{fig:eyecatcher}
    \emph{Event-based continuous color video decompression} uses an initial frame and subsequent events to generate frames.
    The prediction relies on continuous motion estimation, neural synthesis and image generation modules.
    }
    \vspace{-2ex}
\end{figure}

High frame-rate videos (i.e., temporally continuous) are highly desirable for computer vision tasks because they allow an algorithm to avoid problems such as temporal aliasing and motion discontinuities. For example, correspondence becomes much easier if the temporal baseline between two frames is infinitely small. 
However, it is usually infeasible to acquire high-quality frames at high temporal resolution. 
The bandwidth requirement for high frame-rate videos grows proportionally to the frame rate. High-speed cameras often have hardware buffers that cache frames because large amounts of video data cannot be transferred in real time given the limited bandwidth of modern camera output interfaces. The root for this high bandwidth requirement can be attributed to the ubiquitous temporal redundancy in high-speed data. Modern camera sensors are designed to be synchronous, meaning that redundant information is given the same importance as more informative changes in pixels. Furthermore, the global shutter time of frame-based cameras assumes equal exposure of the entire frame, resulting in a limited dynamic range. These hardware limitations significantly increase the difficulties of capturing high-quality continuous videos.

A common approach to produce such videos is to learn motion interpolation networks that upsample low-frame--rate videos by predicting intermediate frames. 
However, the interpolation task is inherently ambiguous because many solutions exist between two sparsely sampled frames. 
State-of-the-art approaches produce only plausible middle frames based on hallucinated motions, which are usually assumed to be linear. 
These methods may not reconstruct physically (i.e., geometrically) accurate frames and suffer from common problems in upsampling, such as aliasing. 
Reconstructing visually accurate frames is difficult to solve simply with larger networks and more training data. 
This problem is illustrated in \cref{fig:timeanalysis}.

We propose a novel solution to encoding high-speed video data by equipping the image sensor with a biologically inspired event camera (\cref{fig:eyecatcher}). 
An event camera encodes the changes in log image intensity, outputting a stream of binary events that can be seen as a compressed representation of the image changes. 
Due to the sparse nature of events, the bandwidth requirement is significantly lower compared to traditional image sensors operating at the same rate. 
These characteristics make event sensors ideal for capturing the subtle changes between frames.
The question we approach is: \emph{what is the most effective way to unpack a video from static frames and dynamic events?}

Brandli et al.~\cite{Brandli14iscas} pioneered the research of high-speed video decompression by decoding videos from events and grayscale frames of a DAVIS camera \cite{Brandli14ssc} through direct temporal integration. 
However, direct integration suffers from noise accumulation and produces artifacts caused by discretization. 
Recent works have used events to aid frame-based interpolation~\cite{Tulyakov21cvpr,Tulyakov22cvpr}. 
However, these approaches require pairs of sharp and well-exposed frames, which are susceptible to sudden degradation caused by aggressive motions or lighting changes. 
For example, if the camera experiences a sudden drop or observes a sudden fast motion, the frames will be corrupted, yielding a blurry interpolated frame. 
Additionally, since interpolation needs to wait for the next frame, the prediction latency is high.

Building upon previous methodologies, we formulate the task of \emph{event-based continuous color video decompression}. 
In this task, given an initial color image and an aligned event stream, the goal is to recover high-quality color images at any query timestamps within the event stream (\cref{fig:eyecatcher}). 
Our approach features two ways that encode long-term continuous videos. 
First, we use a neural synthesis module that factorizes the continuous spatiotemporal feature into three feature planes. 
The design significantly reduces the computational burden of event voxelization. 
Then we introduce a continuous trajectory field module that parameterizes dense pixel trajectories with motion priors. 
Both branches merge into a multiscale feature fusion network that flexibly generates color images at any desired timestamp.
Additionally, we develop an open-source hardware-synchronized single-lens beam splitter for more precise data acquisition, which can facilitate the creation of hybrid image-event datasets. 
We used it to record a novel dataset tailored for the continuous color video decompression task. 
In addition to a photometric evaluation, we use the decompressed video in challenging downstream tasks such as Gaussian Splatting 3D reconstruction and camera fiducial tag detection, even in difficult lighting and motion conditions. 
Our contributions are summarized as follows:
\begin{itemize}
    \item We formulate the task of event-based continuous color video decompression from a single frame, a challenging type of event-based video prediction with minimal input.
    \item We present a novel approach to solve the task via a joint synthesis and motion estimation pipeline. 
    The synthesis module %
    encodes event-based spatiotemporal features. 
    The motion estimation module computes time-continuous nonlinear trajectories parameterized by learned priors.
    \item We evaluate against various image- and event-based baselines, showing state-of-the-art performance. 
    In video decompression, we outperform baseline methods by 3.61 dB in PSNR and 33\% in LPIPS. 
    In the downstream tasks, our decompressed method increases AprilTag detection by 20\% and produces sharper Gaussian Splatting models.
    \item We contribute a novel event dataset using a meticulously designed beam-splitter setup with a shared-lens design.     
\end{itemize}

\begin{figure}[tb]
    \centering
    \includegraphics[clip, trim={12cm, 6.9cm, 5.5cm, 7.57cm}, width=\linewidth]{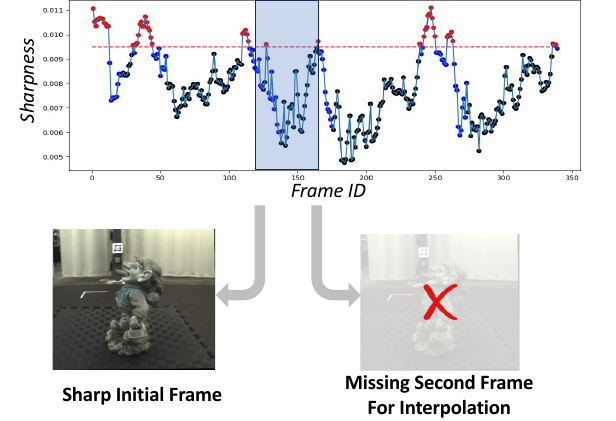}
    \vspace{-2ex}
    \caption{\label{fig:timeanalysis}
    For natural camera motions, it is common to have sharp frames followed by blurry frames (shaded blue region above), which prohibits interpolation methods. 
    Our method is able to reconstruct in these scenarios due to the removal of the dependency on the second frame. 
    }
    \vspace{-2ex}
\end{figure}

\section{Related Work}
\subsection{Video Decoding from Images}
\textbf{Video Frame interpolation (VFI)} methods focus on inserting intermediate frames by estimating motion information in low-frame-rate videos. 
Cheng et al.~\cite{cheng2020video} adaptively learn separable convolution filters to sample more information pixels.
FILM~\cite{reda2022film} warps a multi-scale feature pyramid to enable interpolation with large motion.
FLAVR~\cite{kalluri2023flavr} uses spatiotemporal kernels to replace warping operations in flow-based VFI methods. AdaCof~\cite{lee2020adacof} combines kernel-based and flow-based modules to collaboratively predict interpolated frames. 
Niklaus et al.~\cite{niklaus2023splatting} proposed the use of Softmax splatting to replace image warping to achieve higher quality in occluded regions. 
Recently, TimeLens~\cite{Tulyakov21cvpr} and TimeLens++~\cite{Tulyakov22cvpr} enhanced VFI methods by introducing a colocated event camera. 
More recently, CBMNet~\cite{kim2023event} introduced interactive Attention-based blocks to improve the performance while costing more computation. TLXNet+~\cite{ma2025timelens} recursively recovers the motion trajectories of pixels at small steps, which improves both the warping performance and computational speed. However, interpolation approaches introduce latency and are susceptible to sudden large motions and lighting variations. 
Our proposed method addresses these issues by eliminating the dependency on a second frame. Additionally, our approach encodes long-range motion rather than the small motion typically found between two consecutive frames.

\textbf{Video prediction} methods use previous frames to predict future frames. Due to the missing second frame, the dynamics of the scene need to be modeled and extrapolated into the future. 
Xue et al.~\cite{xue2016visual} adopted a probabilistic framework for future frame synthesis, allowing multiple possible future frames to be generated. 
Lotter et al.~\cite{lotter2016deep} used Deep Predictive Coding Networks (DPCNs) to learn the structure of video data without manual annotation. 
Liu et al.~\cite{liu2017video} introduced a dense volumetric flow representation that produced coherent videos. 
Lee et al.~\cite{lee2018stochastic} proposed using latent variational variable models and a generative adversarial framework to achieve diversity and quality. 
More recently, DMVFN~\cite{hu2023cvpr} used a differentiable routing module to perceive the motion scales of a video. 
Due to the increasing quality of generative models, a new line of research arises for unguided video synthesis from a single image~\cite{hao2018controllable,holynski2021animating,li2023generative,endo2019animating}. 
These methods do not focus on accurate geometry, but rather on perceptual quality and diversity.

\subsection{Event-based Motion Estimation}

Event data have been shown to be suitable for fine motion estimation due to their high temporal resolution and relative invariance to light changes~\cite{Benosman12nn, Orchard13biocas, Brosch15fns, Gallego18cvpr, Zhu19cvpr, Ye19arxiv, Gehrig21threedv, wang2022ev, wang2025evimo, Hamann24eccv, wang2024continuous}. 
Early work focused on the computation of asynchronous optical flow based on approximating derivatives~\cite{Benosman12nn} or fitting spatiotemporal planes~\cite{Benosman14tnnls}. 
Subsequently, data-driven techniques have shown to enhance the robustness of flow computation in the presence of event noise. 
Zhu et al.~\cite{Zhu18rss} proposed a self-supervised method for learning flow by warping consecutive image pairs and measuring photometric consistency. 
Later, the approach became unsupervised~\cite{Zhu19cvpr} by formulating the loss as maximizing the contrast of flow-warped events~\cite{Gallego18cvpr}. 
Contrast maximization has been profitable for other tasks, such as 
ego-motion estimation~\cite{Gallego19cvpr,Zhu19cvpr,Guo24tro}, 
depth estimation~\cite{Zhu19cvpr,Zhu18eccv,Ghosh22aisy} and 
motion segmentation~\cite{Stoffregen19iccv,Zhou21tnnls}. 
Optimizing contrast loss directly is challenging due to event collapse, and therefore recent literature focuses on ``taming'' this loss function with some regularization~\cite{Shiba22aisy,Shiba24pami,Paredes23iccv}. 
Recent studies have also improved network architectures~\cite{Gehrig21threedv,Hamann24eccv} and proposed motion-equivariant~\cite{zhu2019motion} or motion-robust~\cite{zhang2026match} event representations. 
However, the proper representations to model long-range motion remain understudied. 
Gehrig et al.~\cite{Gehrig24pami} and Tulyakov et al.~\cite{Tulyakov22cvpr} use B-splines to parameterize the trajectory of points. 
Both methods require intermediate color images as input to the network in addition to events to improve motion estimation via photometric consistency.
In contrast, we estimate motion solely from the event stream.

\subsection{Event-only Video Reconstruction}
Reconstructing intensity information from event data has been addressed through two primary methodologies: filtering and learning-based techniques. 
Initial efforts concentrated on developing suitable filters for sparse event signals. 
These filtering methods typically involved classical strategies such as time-integration~\cite{Scheerlinck18accv} or Poisson reconstruction of intermediaries like spatial gradients~\cite{Kim14bmvc}. 
However, these approaches are often plagued by noise and event leakage at corners, and may not produce realistic images \cite{Munda18ijcv}.

Learning-based methods have demonstrated the ability to overcome many of these issues by embedding prior information into image reconstruction models~\cite{Rebecq19pami,weng2021event}, trained on real~\cite{Chaney23cvprw} or synthetic events~\cite{zhu2021eventgan,shi2026evis}. 
Furthermore, these approaches have been extended to incorporate images into the input data stream, enhancing the production of high dynamic range (HDR) images \cite{yang2023learning}. 
Efficient image generation is also achieved through shallow networks, through either recurrent networks \cite{Scheerlinck20wacv} or intermediary networks coupled with traditional processing techniques \cite{Duwek21cvprw,Zhang22pami}. 
These methods cannot reconstruct realistic color videos from grayscale event cameras. Post-processing coloring networks introduces unrealistic colors due to hallucination. In contrast, this work aims to reconstruct high-quality color videos by injecting color information into gray events from easily obtainable high-quality color frames.

\textbf{Concurrent Work.} Continuous space-time decoding~\cite{lu2024hr} and single-frame video rewinding~\cite{chen2024timerewind} were proposed concurrently. Our arXiv preprint~\cite{wang2023event} was made publicly available prior to~\cite{chen2024timerewind} and~\cite{lu2024hr}. While related,~\cite{chen2024timerewind} does not address continuous-time decoding, and~\cite{lu2024hr} relies on both start and end frames for spatiotemporal upsampling.

\begin{figure*}[t]
    \centering
    \includegraphics[trim=2cm 5.6cm 5.5cm 6.5cm,clip,width=\linewidth]{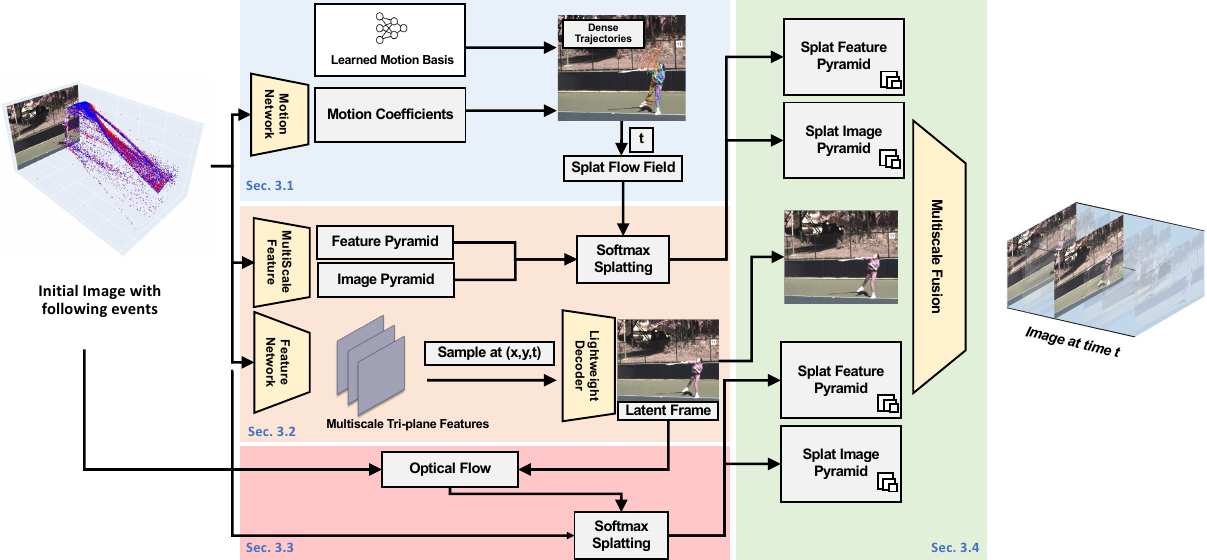}
    \caption{\label{fig:pipeline}\emph{Overview}. 
    The initial frame and the long range event volumes are concatenated forming the network input.
    \textbf{Top blue box (\cref{sec:event_flow})}: A continuous motion network regresses the motion coefficients for generating the point trajectory for every pixel from events and the initial frame.
    \textbf{Bottom orange box (\cref{sec:synth})}: The input is projected to tri-plane features. A lightweight decoder queries the features and synthesizes pixel RGB values. 
    \textbf{Bottom red box (\cref{sec:latent_flow})}: Optical flow is compuated between the intial frame and the synthesized latent frame. We compute another set of features and warped images as pyramids.
    \textbf{Right green box (\cref{sec:method:multiscalefusion})}: Finally, the splatted features and images are merged with the synthesized images via a mult-scale fusion network into a high-quality color image prediction.
    }
\end{figure*}

\begin{figure}[t]
    \centering
    {\includegraphics[trim={0 0 17.5cm 0},clip,width=.96\linewidth]{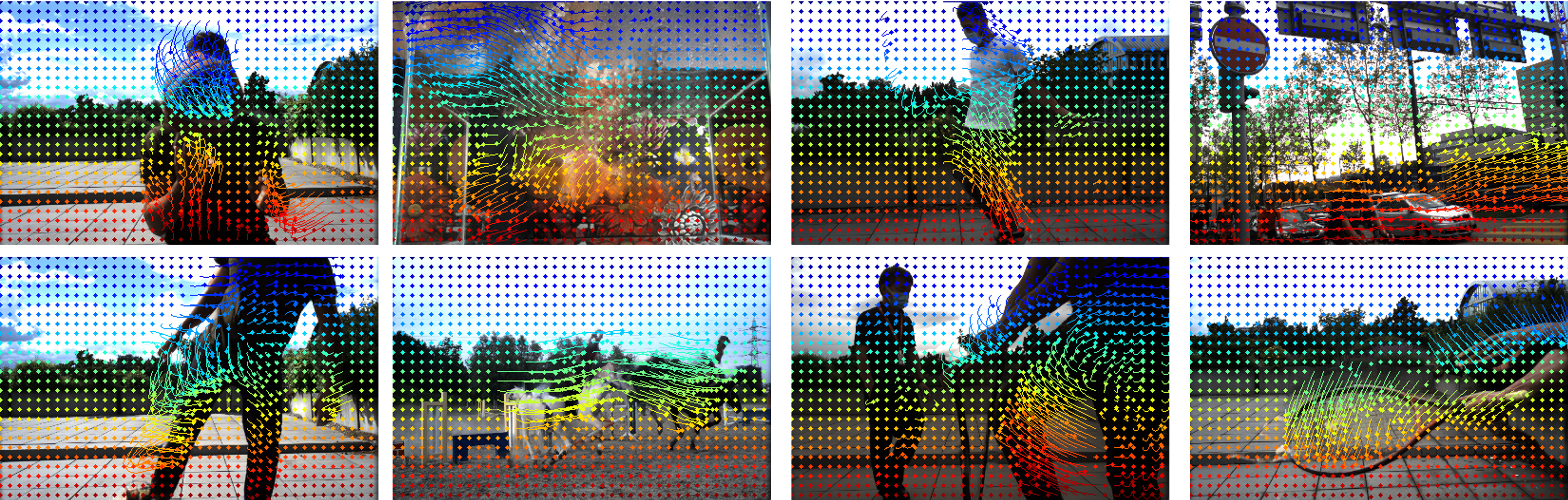}}
    \caption{\label{fig:traj}
    Continuous long-term trajectory output on test sequences of BS-ERGB~\cite{Tulyakov22cvpr} dataset. 
    We show pixel tracks of uniformly initialized features using motion coefficients predicted from events. 
    The network outputs dense tracks (i.e., per-pixel) in a single feedforward pass. 
    Our continuous basis-enabled motion module can decode complex long-range motions up to 1 second.
    }
    \vspace{-1em}    
\end{figure}

\section{Method}
\def\figWidth{0.2\linewidth}
\begin{figure*}[t]
	\centering
    {\scriptsize
    \setlength{\tabcolsep}{2pt}
	\begin{tabular}{
	>{\centering\arraybackslash}m{0.3cm} 
	>{\centering\arraybackslash}m{\figWidth} 
	>{\centering\arraybackslash}m{\figWidth} 
	>{\centering\arraybackslash}m{\figWidth} 
	>{\centering\arraybackslash}m{\figWidth}}

        \rotatebox{90}{\makecell{Original}}
		&\includegraphics[clip,trim={2.7cm 5.4cm 0.3cm 0.3cm},width=\linewidth]{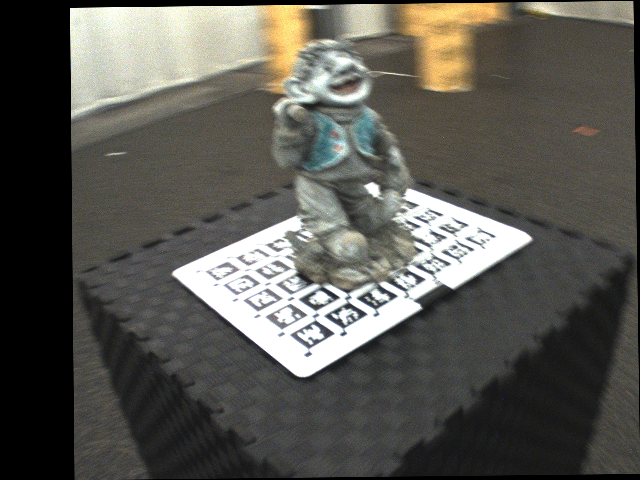}
		&\includegraphics[clip,trim={2.7cm 5.4cm 0.3cm 0.3cm},width=\linewidth]{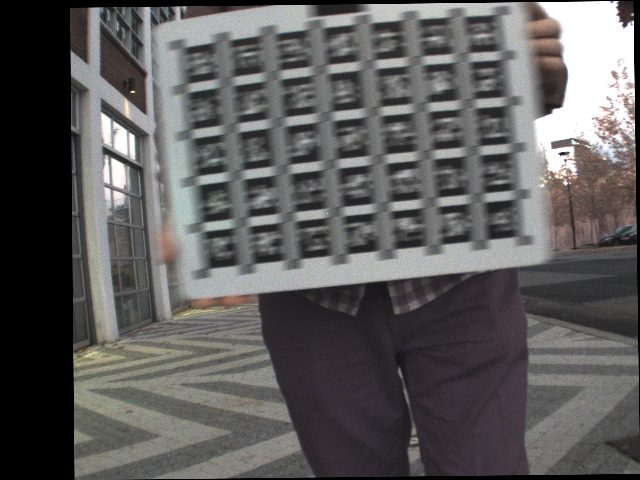}
		&\includegraphics[clip,trim={0cm 2cm 0cm 0cm},width=\linewidth]{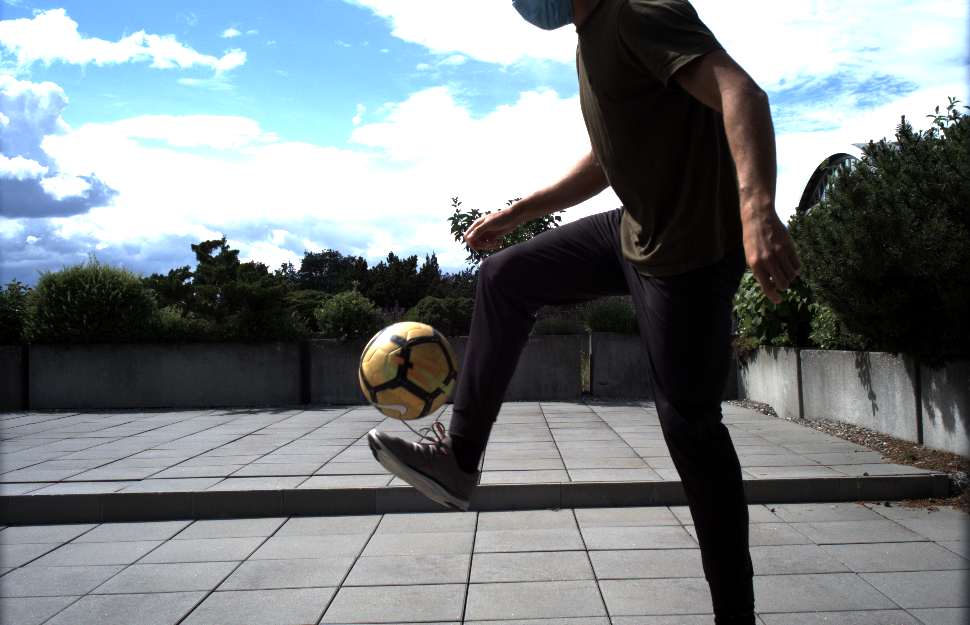}
		&\includegraphics[clip,trim={0cm 2cm 0cm 0cm},width=\linewidth]{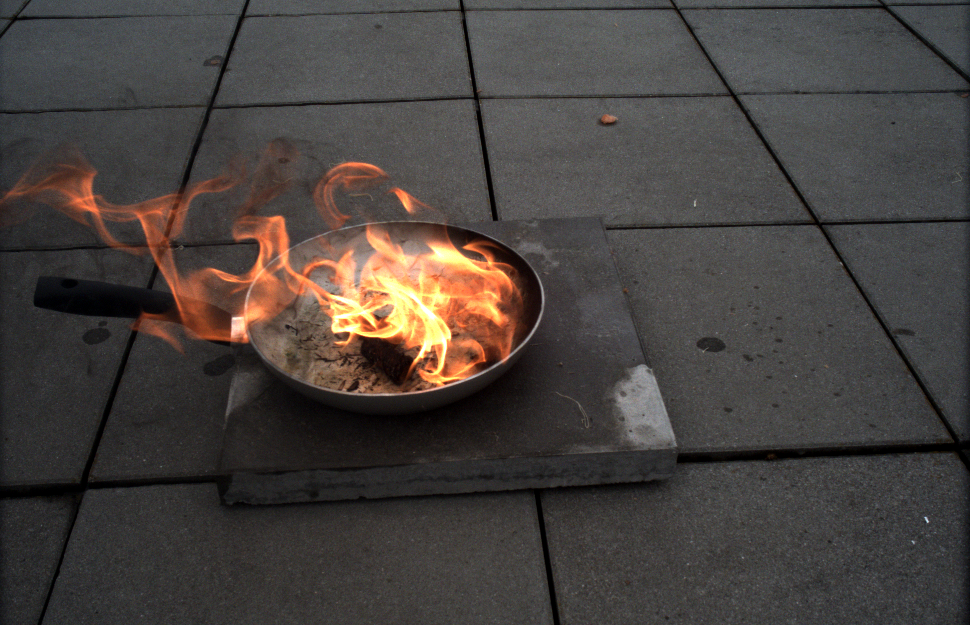}
        \\

        \rotatebox{90}{\makecell{Ours}}
		&\includegraphics[clip,trim={2.7cm 5.4cm 0.3cm 0.3cm},width=\linewidth]{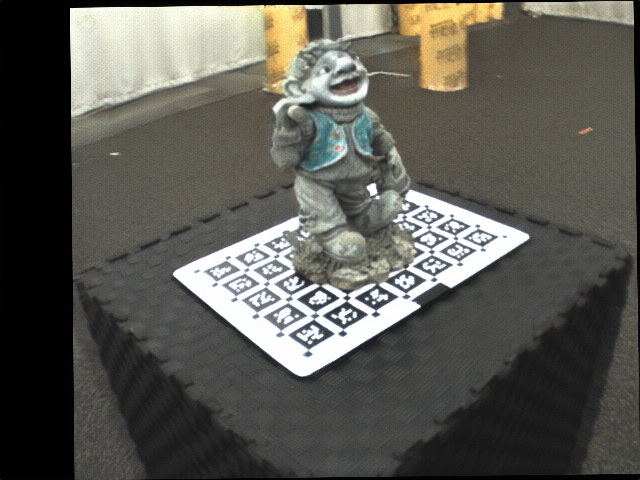}
		&\includegraphics[clip,trim={2.7cm 5.4cm 0.3cm 0.3cm},width=\linewidth]{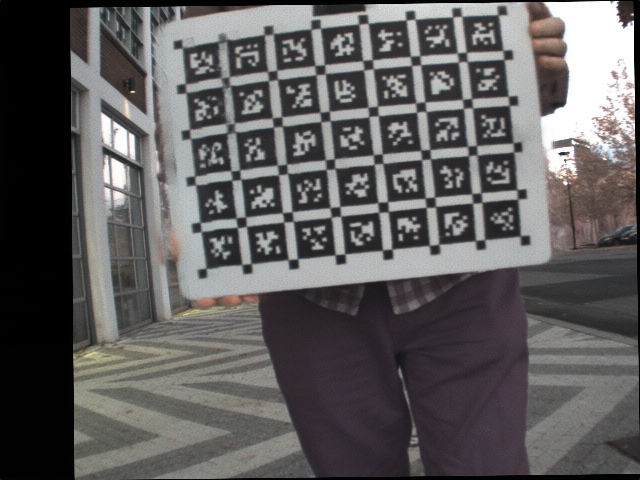}
		&\includegraphics[clip,trim={0cm 2cm 0cm 0cm},width=\linewidth]{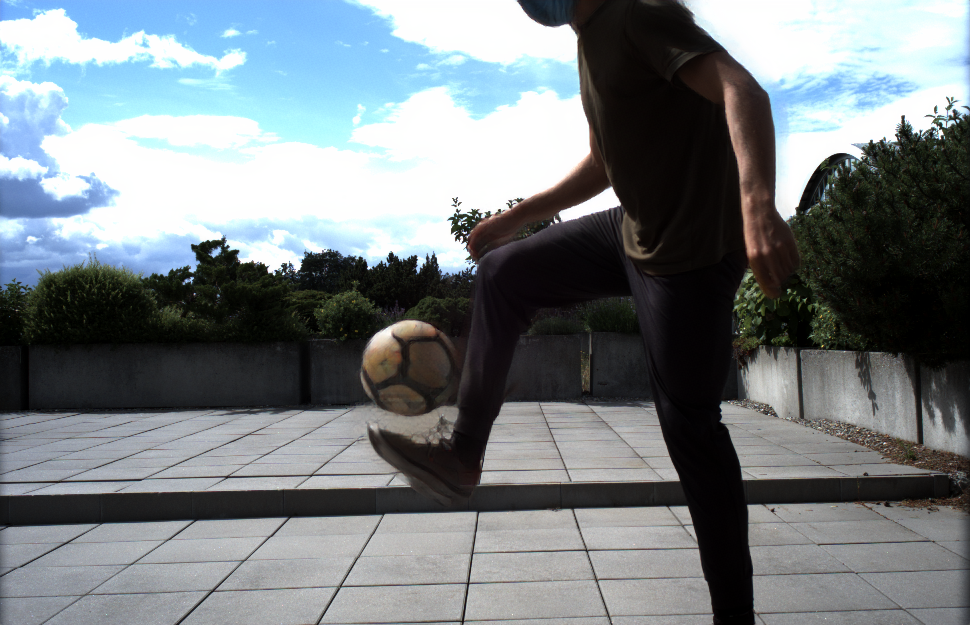}
		&\includegraphics[clip,trim={0cm 2cm 0cm 0cm},width=\linewidth]{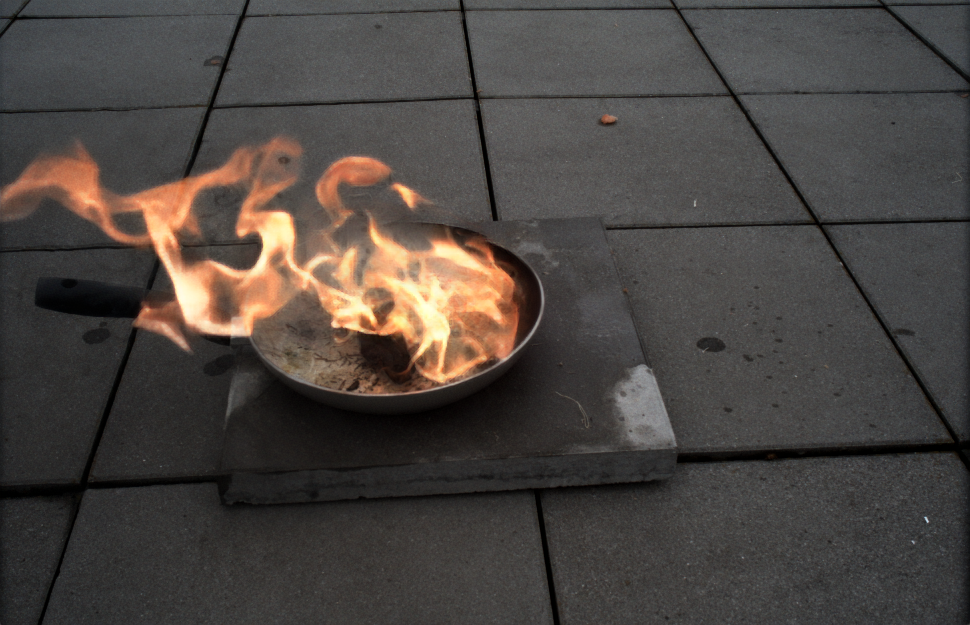}
        \\

        \rotatebox{90}{\makecell{FILM~\cite{reda2022film}}}
		&\includegraphics[clip,trim={2.7cm 5.4cm 0.3cm 0.3cm},width=\linewidth]{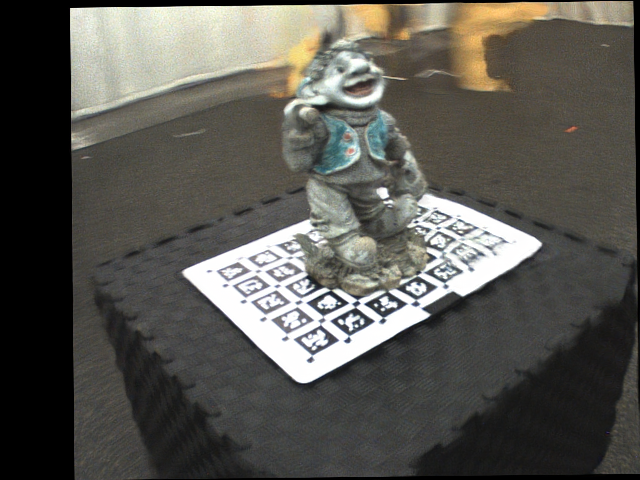}
		&\includegraphics[clip,trim={2.7cm 5.4cm 0.3cm 0.3cm},width=\linewidth]{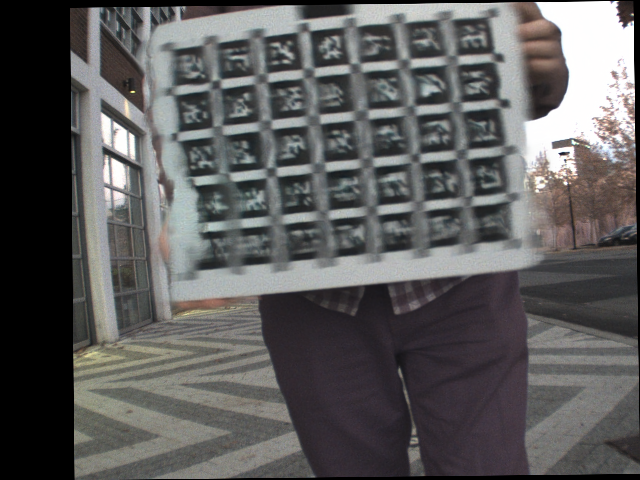}
		&\includegraphics[clip,trim={0cm 2cm 0cm 0cm},width=\linewidth]{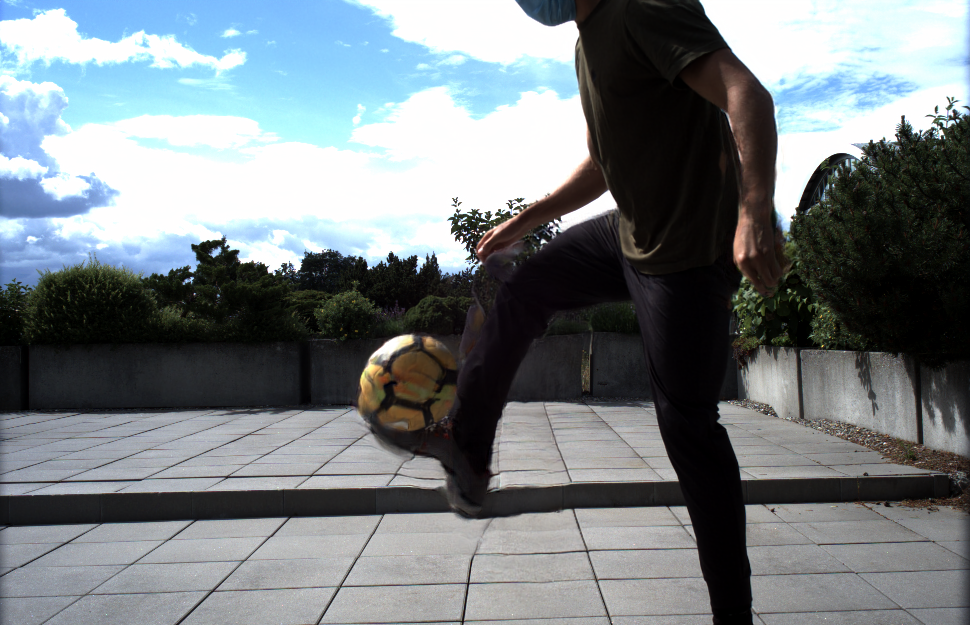}
		&\includegraphics[clip,trim={0cm 2cm 0cm 0cm},width=\linewidth]{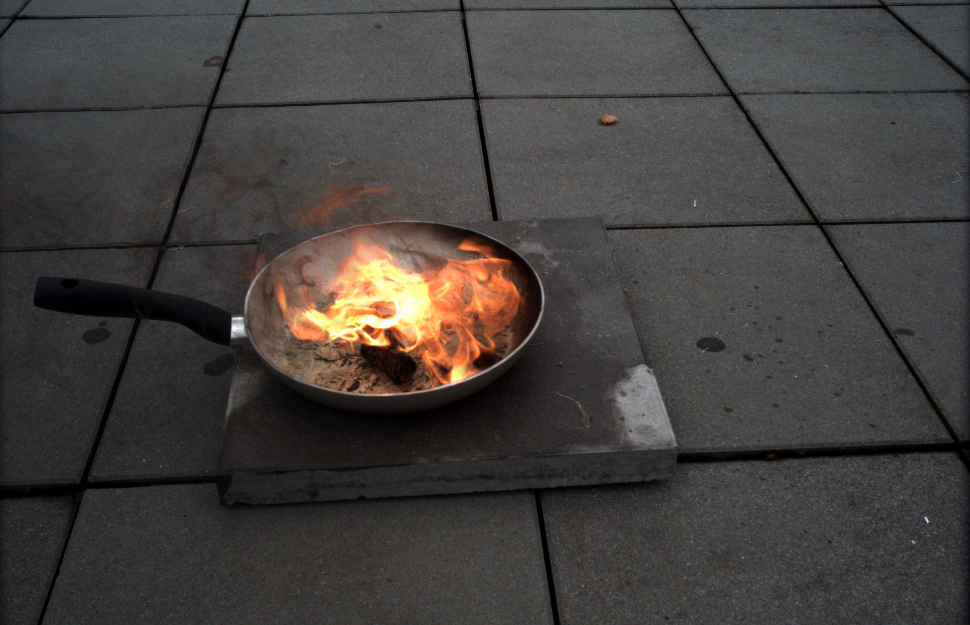}
        \\

        \rotatebox{90}{\makecell{DMVFN~\cite{hu2023cvpr}}}
		&\includegraphics[clip,trim={2.7cm 5.4cm 0.3cm 0.3cm},width=\linewidth]{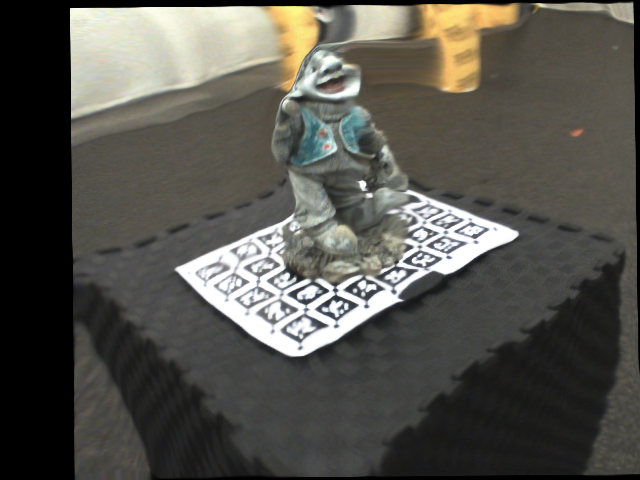}
		&\includegraphics[clip,trim={2.7cm 5.4cm 0.3cm 0.3cm},width=\linewidth]{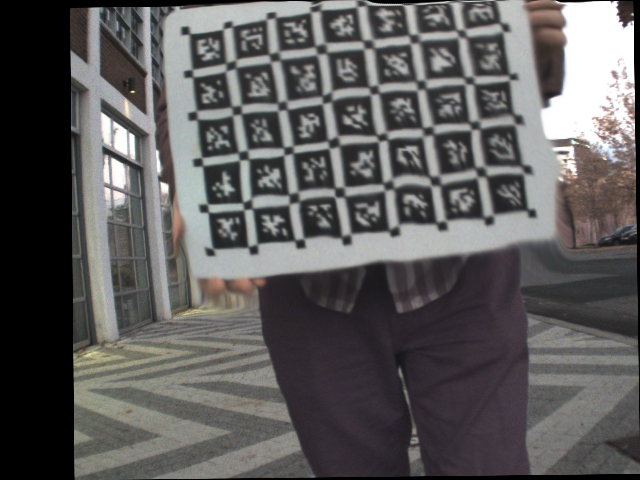}
		&\includegraphics[clip,trim={0cm 2cm 0cm 0cm},width=\linewidth]{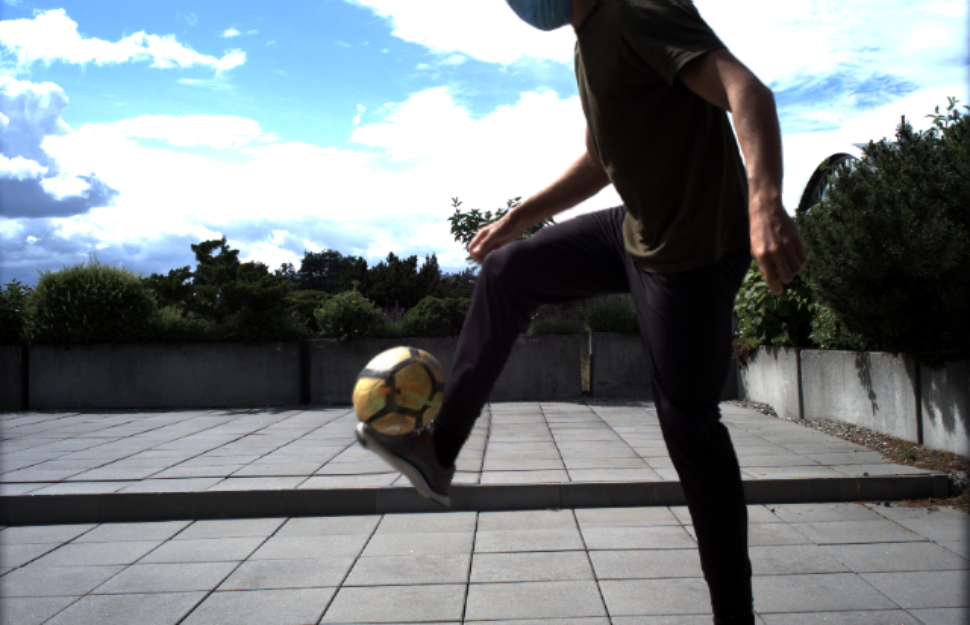}
		&\includegraphics[clip,trim={0cm 2cm 0cm 0cm},width=\linewidth]{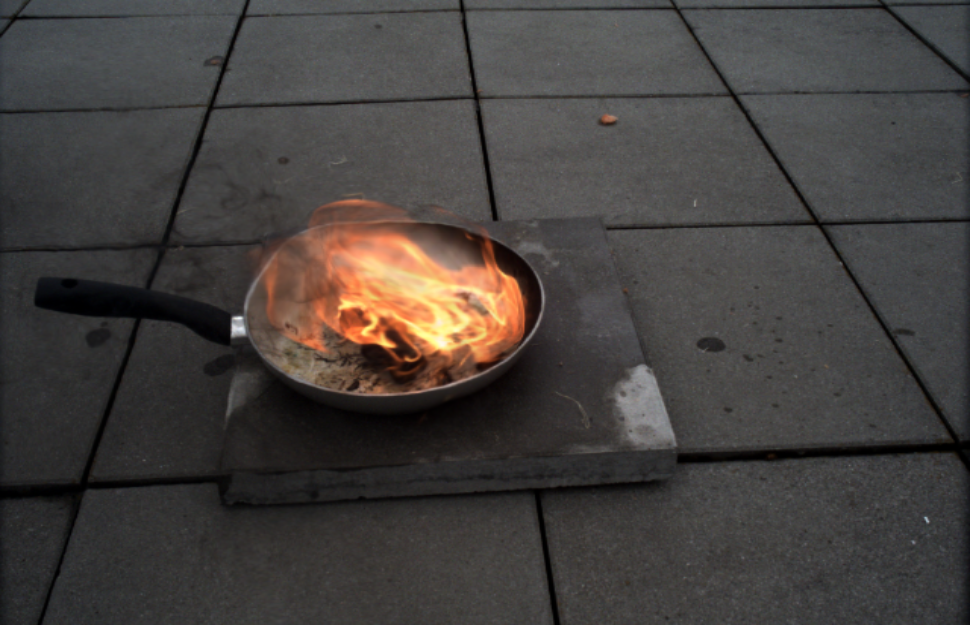}
        \\
        
		& \textbf{(a)} Gnome
		& \textbf{(b)} Aprilgrid
		& \textbf{(c)} Football
		& \textbf{(d)} Fire
	\end{tabular}
	}
	\caption{\label{fig:exp:qualitative}
 Qualitative Evaluation: We present two qualitative examples from \ourdata{} and BS-ERGB~\cite{Tulyakov22cvpr}, respectively. Our method, \ours{}, demonstrates enhanced accuracy in reconstructing geometry, even with challenging deformable subjects, such as in the ``Fire'' sequence (d). 
 This improvement is attributed to the effective use of event data. Notably, in low-light conditions, as seen in the ``Gnome'' sequence (a), our approach markedly reduces motion blur compared to traditional image acquisition methods. While FILM~\cite{reda2022film} generates plausible results, it fails to accurately predict geometry in all examples. DMVFN~\cite{hu2023cvpr} struggles with occlusions, particularly those caused by rotational movements, as evident in the ``Gnome'' sequence.
	}
\end{figure*}

\Cref{fig:pipeline} shows an overview of our approach.
\Cref{sec:event_flow} presents the motion module to produce a set of continuous point trajectories.
\Cref{sec:synth} describes how the event-based neural integration model synthesizes a rough reconstruction of a target frame.
\Cref{sec:latent_flow} states how we refine the features by using image-based flow computed between latent frames.
These three intermediate results are combined in a multiscale feature fusion network (\cref{sec:method:multiscalefusion}). 
Finally, the loss functions are specified in \cref{sec:method:lossfunctions}.

\subsection{Continuous Event-based Trajectory Field}
\label{sec:event_flow}
To model long-range motions, one needs to carefully choose the motion representation. 
Naively, flow defined at discrete timestamps can be directly regressed using a motion network. 
However, the problem with this formulation is that the motion is completely unconstrained and does not exploit the temporal smoothness of the pixel trajectories. 

Motion priors in terms of motion basis have been proposed to address this issue in 3D dynamic motion modeling~\cite{wang2021neural,li2023dynibar,li2023generative}. 
The motion trajectories of a set of point seeds $\{\bx_i(0)=\bu_i\}_{i=1}^{\numPoints}$ defined at the beginning of the video, with $\bu=(x,y)^\top$, 
over a set of frame timestamps $\{t_j\}_{j=1}^{\numTimestamps}$ can be modeled as 
\begin{equation}
\label{eq:pointtrajectories:discretetime}
    \textstyle 
    \bx_i(t_j) = \sum_{k=1}^{\numBases} \alpha_k(\bu_i) \theta_{k,j}.
\end{equation}
where $\alpha_k(\bu_i)$ are motion coefficients (per seed point) 
with respect to basis values (at frame timestamps) $\theta_{k,j}=:\mB$. %
Here, $\mB$ is a set of shared parameters that do not depend on the input. 
However, this formulation does not allow for querying arbitrary time, as $\mB$ is fix-sized. 

Events are quasi-continuous in nature, hence we propose representing their trajectory field using a continuous-time function~\cite{wang2021neural}.
We replace $\mB$ with a set of $\numBases$ learned basis functions $\{g^{\theta}_k(t)\}_{k=1}^{\numBases}$, $t \in \Real$. 
This continuous formulation allows the network to map the quasi-continuous motion information of events to continuous trajectories. 
Specifically, the trajectory corresponding to a seed point $\bu_i$ is given by 
\begin{equation}
    \textstyle 
    \bx_i(t) = \sum_{k=1}^{\numBases} \alpha_k(\bu_i) g^{\theta}_k(t),
\end{equation}
where the two main changes with respect to \eqref{eq:pointtrajectories:discretetime} are that the basis changed from discrete-time to continuous-time and the query time can be any $t$, not just the finite set $\{t_j\}$.

The selection of a motion basis offers several possibilities, including the Discrete Cosine Transform~\cite{li2023dynibar,wang2021neural}, Fourier basis~\cite{li2023generative}, and polynomial basis. 
In our experiments, the hand-crafted basis tends to be dependent on carefully chosen hyperparameters, such as the frequency band. 
To address this, we employ a learned multi-layer perceptron (MLP) to model the motion basis and optimize it during training 
(Motion Network in \cref{fig:pipeline}). 
The number of basis functions is correlated with the motion complexity of the video. 
For datasets that we primarily study, we use ${\numBases}=5$. %

\Cref{fig:traj} shows sample trajectories predicted by our method on the BS-ERGB dataset~\cite{Tulyakov22cvpr}; 
our method captures complex non-linear motion over a long-range time window.

\subsection{Event Features Encoding for Neural Synthesis}
\label{sec:synth}
Reconstructing continuous videos requires a compact feature space of the events, so that individual frames can be synthesized without significantly increasing the memory requirement. In particular, the fine temporal information in events needs to be handled carefully to avoid aliasing.

\textbf{Event-based Tri-Planes Feature Encoding.} 
Encoding a continuous video requires the ability to query an image at a high temporal resolution. To make high-speed video generation computationally feasible, the synthesis operation should be done once for each video, and the sampling operation at each step should be comparatively cheap. Therefore, we adopt a Tri-plane parameterization for a reconstructed video. 
Specifically, we assume that the characteristics are encoded on three orthogonal planes, $x$-$y$, $x$-$t$, $y$-$t$, whose features are denoted by $\featplane_{xy}$, $\featplane_{xt}$ and $\featplane_{yt}$. 
These three multi-channel images contain the feature information for the continuous video field on discretized grids. 
For dense video, we ensure that the resolution of the time dimension in $\featplane_{xt}$ and $\featplane_{yt}$ is sufficiently large for fine-grained temporal information. This significantly reduces the feature dimension from three-dimensional to two-dimensional. 

We use a multiscale feature extractor to directly regress $\featplane_{xy}$, $\featplane_{xt}$ and $\featplane_{yt}$ at three spatial scales. 
Following K-Planes NeRF~\cite{fridovich2023k}, we use Hadamard product to fuse the multi-channel image features at each scale.

\textbf{Decoder}. 
A lightweight decoder $\phi$ is used to decode the three feature vectors into an RGB value $\hat{I}$. 
Formally, given a spatiotemporal coordinate $\bq = (x, y, \tau)^\top$, the decoded image value can be written as:
\begin{equation}
\label{eq:synthesizedrgb}
\hat{I}_\tau(x,y) \!=\!\! \phi \bigl( \featplane_{xy} (\pi_{xy}(\bq)), \featplane_{xt} (\pi_{xt}(\bq)), \featplane_{yt} (\pi_{yt}(\bq)) \bigr)
\end{equation}
where $\pi_{xy}:\Real^3\to\Real^2$ denotes the operation that maps $\bq$ to its corresponding position in the $x$-$y$ plane (and similarly for the other two projections). 

A key benefit of this parameterization is the shift of computational cost to the encoder, which reduces the cost of high FPS inference. Our synthesis encoder runs once to reconstruct a short video clip, and the lightweight decoder can predict in parallel with little computational cost. 
Moreover, unlike the synthesis model in \cite{Tulyakov22cvpr}, we no longer need to build a different event volume for every inference. 

\subsection{Latent-Frame Flow Refinement}
\label{sec:latent_flow}
The continuous flow field in \cref{sec:event_flow} captures the long-range motion within events. However, there is significant noise in the event-based flow field caused by noisy measurements and areas that do not have enough contrast to generate events. 
Therefore, it is critical to learn the grouping of the flow field for spatially consistent warping. To this end, we propose a novel latent frame model that takes advantage of the iterative matching power of frame-based flow networks.

In \cref{sec:synth}, we described how an intermediate latent frame $\hat{I}_t$ could be obtained through our neural event integration module \eqref{eq:synthesizedrgb}. 
We use this latent frame for computing the latent-frame flow via iterative flow refinement using RAFT~\cite{Teed20eccv}. 
Given a latent frame $\hat{I}(t)$ and an initial frame $I_0$, we build a correlation pyramid based on the features of the images, 
$\f_t \doteq F (\hat{I}_t)$ and $\f_0 \doteq F (\hat{I}_0)$.
Following RAFT, we produce correlation volumes $C$ at each step:
\begin{equation}
C \bigl( \f_t(i,j), \f_0(k, l) \bigr) = (\f_t(i,j))^\top \f_0(k, l)
\label{eqn:corr}
\vspace{-1ex}
\end{equation}
where $i,j,k,l$ are the spatial coordinates of the features.
At each of its 12 steps $m$, RAFT takes the images and the correlation volume at each scale and iteratively produces an update $\delta_{m}(t)$ to the previous displacement $\Delta_{m-1}(t)$. 
The flow prediction at step $m$ is $\Delta_{m}(t) = \Delta_{m-1}(t) + \delta_m(t).$

The correlation volume in \eqref{eqn:corr} resembles a matching process without a motion model, allowing matching at any two arbitrary times. 
A key factor that motivates our choice for this latent model is the empirical observation that the correlation function \eqref{eqn:corr} %
 is robust to color changes and noisy images. 
While the neural synthesis module in \cref{sec:synth} produces images that are inaccurate in color due to missing information, 
they show enough texture for RAFT to build meaningful matching correlation modules. 
To reduce the training burden, we directly use the RAFT weights pre-trained on FlyingChairs~\cite{DFIB15} and FlyingThings3D~\cite{MIFDB16}.

\subsection{Multi-scale Feature Fusion}
\label{sec:method:multiscalefusion}
The three main outputs of the intermediate networks in \cref{sec:event_flow,sec:synth,sec:latent_flow} are fused via a multiscale feature fusion network.
For the sake of simplicity, we remove the subscript of the point index in all symbols. 
We denote $M_t$ as the displacement of a point at time $t$, $\Tilde{M}_t$ the latent displacement, $\Psi_0$ a feature pyramid computed from the starting frame $I_0$, 
$\cI_0$ an image pyramid computed from $I_0$,
and $\hat{\cI}_t$ an image pyramid computed from $\hat{I}_t$.
In addition, we define the forward splatting function $\mathcal{T}(\cdot)$, 
and $\mathcal{C}(\cdot)$ as pyramid concatenation. 
We splat multiscale features and image pyramids with the two flow fields. 
The input to the fusion model is 
\begin{equation}
\mathcal{G} \doteq \mathcal{C} \bigl(\mathcal{T}_{M_t}(\Psi_0), \mathcal{T}_{M_t} (\cI_0), \mathcal{T}_{\Tilde{M}_t}(\cI_0), \mathcal{T}_{\Tilde{M}_t}(\Psi_0), \hat{\cI}_t)\bigr).
\vspace{-1ex}
\end{equation}
The final image prediction is passed through the multi-level merging network $f_m$, implemented by a series of convolution layers with small receptive field and non-linear activation, to produce the final image prediction $I_t \doteq f_{m} (\mathcal{G})$. 

We use Softmax Splatting~\cite{niklaus2020softmax} to warp images and features at each pyramid scale, with learned multiscale splatting weights predicted along with motion parameters. The splatting shows instability for flow supervision, so the gradient of the flow input to the splatting operation is stopped.

\subsection{Training}
\label{sec:method:lossfunctions}
\textbf{Optical Flow Loss}. 
We use two approaches for supervising the predicted continuous optical flow field: supervised and self-supervised. 
We adopt the image-based warping loss proposed in EV-FlowNet~\cite{Zhu18rss, zhu2021eventgan}. 
This allows the network to learn flow solely based on photometric consistency. 
Given the forward flow displacement computed from the motion network, we bilinearly sample image $I_t$ based on the backward warp field to $t=0$. 
We write the warping operation as 
$\hat{I}_{\text{warp}} = \mathcal{B}_{\Tilde{M}(t)}(I_t)$.
The warping loss comprises a photometric loss $L_{\text{photo}}$ and a smoothness loss $L_{\text{smooth}}$.
\begin{equation}
L_{\text{photo}} (I_0, \hat{I}_{\text{warp}})= \rho (I_0- \hat{I}_{\text{warp}}; \beta)
\end{equation}
where $\rho(\cdot)$ is the Charbonnier loss function $\rho(x) = \sqrt{x^2 + \epsilon^2} $, where $\epsilon$ is a small constant.
The smoothness loss regularizes the flow field to avoid the aperture problem in the classical flow estimation problem. We use second-order smoothness $L_\text{smooth}$, which is the norm of the gradient of the image gradient.

For supervised loss, we compute the L1 loss between the predicted displacement $M_t$ and the RAFT-based flow $W_t$:
\begin{equation}
    L_{\text{flow}}(M_t, W_t) = \|M_t - W_t\|_1
\end{equation}
The supervised loss helps the network learn longer-term consistency since the pseudo ground-truth RAFT flow is computed using correlation-based matching. 
Moreover, it helps the network learn to infer dense flow in places without events, due to the limited contrast sensitivity. 
RAFT flow helps group pixels that have similar motion. 
Although supervised loss provides a direct motion signal, the pre-trained image-based flow suffers from aliasing and missing details. 
Therefore, we use self-supervised flow to help further correct the flow by computing photoconsistency on images.

\textbf{Image Reconstruction Loss}. 
We combine the L1 loss and the Perceptual loss~\cite{johnson2016perceptual} to harness their strengths. 
The L1 loss enhances the accuracy of each pixel, while the Perceptual loss increases the clarity and realism of the result. 
Given reconstruction $I_t$ and image ground truth $I_t^{\text{gt}}$, the losses are:
$L_{1}(I_t, I_t^{\text{gt}}) = \|I_t - I_t^{\text{gt}}\|_1 $ 
and 
$
L_{p}(I_t, I_t^{\text{gt}}) = \frac{1}{J} \sum_{j=1}^{J} \|\Phi_j(I_t) - \Phi_j(I_t^{\text{gt}}) \|_1,
$
where $\Phi_j(\cdot)$ is the deep features extraction operator, utilizing a backbone VGG-19 network \cite{Simonyan15iclr} at level $j$. 
We weigh these two losses equally. 
The compound loss is computed on both the event-based neural synthesis results and the final output. The \textbf{total loss} is a weighted sum of the losses above:
\begin{equation}
    \mathcal{L} = \lambda_{\text{photo}} L_{\text{photo}} 
    + \lambda_{\text{flow}} L_{\text{flow}} 
    + \lambda_\text{reconstr}(L_1 + L_p).
\end{equation}
For training, we use a sequential strategy by training the trajectory field branch and synthesis branch independently first, and then jointly train the system. 
The details of training can be found in the Supplementary Material.

\section{Experiments}
We evaluate the reconstruction quality of our method on BS-ERGB~\cite{Tulyakov22cvpr} and our newly developed dataset \ourdata{}. 
We report Peak-Signal-to-Noise-Raio (PSNR), Learned Perceptual Image Patch Similarity (LPIPS)~\cite{zhang2018lpips}, and Structural Similarity (SSIM) \cite{Wang04tip}.
Moreover, since events are useful when the scene experiences sudden changes, we record additional test scenes with challenging conditions using our curated beam splitter setup. 
In the following, \cref{sec:exp_synth} reports direct photometric results with high-quality images, 
and \cref{sec:downstream} shows how our method can generate blur-free Gaussian Splatting and detect fiducial tags robustly.

\subsection{Datasets, Baselines and Metrics}

\textbf{Datasets}.
We evaluate on BS-ERGB~\cite{Tulyakov22cvpr} and our dataset \ourdata{}. See Supplemental Materials for details of \ourdata{}.
\begin{itemize}
\item \emph{BS-ERGB}  is an event-and-image dataset that uses a beam splitter. 
The dataset was recorded with a Prophesee Gen4 camera paired with a Flir RGB camera running at 28Hz; the resolution is 970$\times$625 px. It contains common scenes in well-lit environments with a static camera.
\item %
\emph{\ourdata{}} is recorded by our novel single-lens beamsplitter system, which maximizes the consistency of projection between the two cameras (VGA-SilkyEvCam and FLIR). 
\ourdata{} contains a diverse set of scenes with various subject rigidity, camera motions, and lighting conditions.
\end{itemize}

\begin{figure}[tb]
    \centering
    {\includegraphics[clip, trim={1.45cm, 9cm, 15.7cm, 9.25cm}, width=\linewidth]{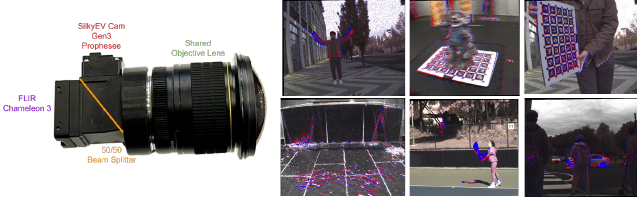}}
    \caption{\label{fig:dataset:ours}We contribute the \ourdata{} dataset for benchmarking this task under challenging light conditions, using our newly designed single-lens beamsplitter.}
    \vspace{-3ex}
\end{figure}

\begin{table}[tb]
    \centering
    \adjustbox{max width=\linewidth}{
    \setlength{\tabcolsep}{6pt}
    \begin{tabular}{lcccc}
    \toprule    
    & Events & PSNR$\uparrow$ & LPIPS$\downarrow$ & SSIM$\uparrow$ \\
    \midrule
    E-RAFT unrolling  & \cmark  & 17.19  & 0.257 & 0.583  \\
    E-RAFT (inpainted) &  \cmark  & 19.58 & 0.260 & 0.623  \\
    RAFT unrolling     & \xmark & 19.35 & 0.197 & 0.629 \\
    RAFT (inpainted)   & \xmark & 20.88 & 0.222 & 0.659 \\
    DMVFN~\cite{hu2023cvpr} & \xmark & 25.93 & 0.111 & 0.767 \\
    Ours              &  \cmark & \bnum{28.68} & \bnum{0.077}  & \bnum{0.802} \\
    \bottomrule
    \end{tabular}
    }
    \vspace{-1ex}
    \caption{\label{tab:penn}Video decompression comparison on our dataset (E2D2). 
    \vspace{-3ex}}
\end{table}

\begin{table}[tb]
    \centering
    \adjustbox{max width=\linewidth}{
    \setlength{\tabcolsep}{4pt}
    \begin{tabular}{lccccc}
    \toprule    
    & Events & \multicolumn{2}{c}{skip 1}& \multicolumn{2}{c}{skip 3}\\
    \cmidrule(l{1mm}r{1mm}){3-4}
    \cmidrule(l{1mm}r{1mm}){5-6}
    &  & PSNR$\uparrow$ & LPIPS$\downarrow$ & PSNR$\uparrow$ & LPIPS$\downarrow$ \\
    \midrule
    DMVFN~\cite{hu2023cvpr} & \xmark & 22.63  & 0.218 & 21.11 & 0.250 \\
    E-RAFT unrolling    & \cmark & 19.28  & 0.171  & 17.49  & 0.257 \\
    E-RAFT (inpainted)   & \cmark & 20.40  & 0.160  & 18.79 & 0.222 \\
    RAFT unrolling      & \xmark & 22.85  & 0.120 & 20.88 & 0.197 \\
    RAFT (inpainted)     & \xmark & 23.41  & 0.097 & 20.88 & 0.142 \\
    Ours                 & \cmark &\textbf{25.40}  & \textbf{0.088} &  \textbf{24.80} & \textbf{0.095}  \\
    \bottomrule
    \end{tabular}
    }
    \vspace{-2ex}
    \caption{\label{tab:bsergb}
    Video decompression on BS-ERGB dataset \cite{Tulyakov22cvpr}. 
    (LPIPS values in italic were reported in \cite{Tulyakov22cvpr}. 
    The metric implementation might differ from ours.)
    \vspace{-2ex}
    }
\end{table}

\textbf{Protocol}.
For quantitative evaluation on \emph{BS-ERGB}~\cite{Tulyakov22cvpr}, we follow the evaluation scheme of TimeLens. 
We take ``keyframes'' that are 1 and 3 frames apart and predict the skipped frames. 
The metrics reported are PSNR, LPIPS \cite{zhang2018lpips} and SSIM \cite{Wang04tip}.
For evaluation on \emph{\ourdata{}}, we use a predicted time of 0.25 seconds rather than a fixed number of frames because the amount of motion for a given scene is correlated to duration and invariant to camera frame-rate.
Since the frame-rate of our dataset ranges from 10 to 66 Hz, this yields 2 to 17 skipped frames. The quantitative test set has 4686 unseen test frames. 

\textbf{Baselines}.
We compare with the following methods:
\begin{itemize}
    \item \emph{Video Prediction}. DMVFN \cite{hu2023cvpr} predicts a frame using its previous two frames. We predict a video iteratively by treating predicted frames as input to the next prediction.
    
    \item \emph{Flow Unrolling}.  In this class of methods, a frame is iteratively warped forward in multiple steps using the backward optical flow of RAFT \cite{Teed20eccv} and E-RAFT \cite{Gehrig21threedv}. %
    
    \item \emph{Inpainted Flow unrolling}. For both flow unrolling methods, we additionally refine the predictions by training a U-Net to reconstruct the original frames from the intermediate (flow unrolling) predictions.
\end{itemize}

\subsection{Video Reconstruction Results}
\label{sec:exp_synth}

The evaluation results on \ourdata{} and BS-ERGB~\cite{Tulyakov22cvpr} datasets are shown in \cref{tab:penn,tab:bsergb}, respectively. 
The tables are vertically divided in two: two-frame interpolation (VFI) and singe-frame decompression.
Two-frame interpolation methods use the keyframe before and after an evaluated frame.
Single-frame decompression methods use only the previous keyframe.
The task is considerably harder as there is no information about occlusion regions in the initial frame. 
Nonetheless, our method outperforms all single-frame decompression methods by up to \textbf{2.7 dB} on \ourdata{} and \textbf{3.61 dB} on BS-ERGB (skip 3) in PSRN.
In LPIPS, \ours{} achieves a \textbf{33\%} decrease on BS-ERGB and \textbf{31\%} decrease on \ourdata{} over the best baseline method.
Our research demonstrates that \ours{} has more advantages when applied to \ourdata{}, as evidenced by the qualitative results in \cref{fig:exp:qualitative}, which demonstrate that interpolation methods are vulnerable to corrupted second frames.

\textbf{Video Frame Interpolation.} 
VFI methods provide an upper bound of our frame interpolation methods when both previous frames and next frames are sharp and well exposed. 
On BS-ERGB dataset, the best performing VFI method is TLXNet+~\cite{ma2025timelens}, which has a PSNR of 29.46 dB (skip 1) and 28.79 dB (skip 3). 
The best performing method on \ourdata{} is FILM~\cite{reda2022film}, which scores a PSNR of 28.16 dB. 
\ours{} performs competitively with both methods, while allowing inference with only one starting frame.

\def\figmethodwidth{.9\linewidth}
\begin{figure}[tb]
    \centering
\begin{subfigure}{\figmethodwidth}
    \centering
    \includegraphics[trim=0cm 4cm 0cm 2cm,clip,width=0.49\linewidth]{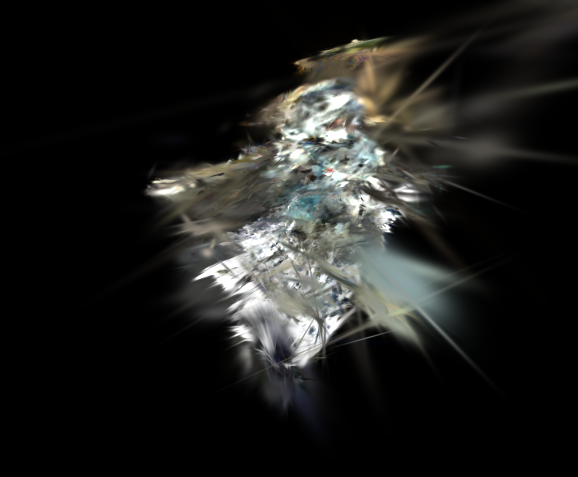}
    \includegraphics[trim=0cm 5.5cm 0cm 1cm,clip,width=0.49\linewidth]{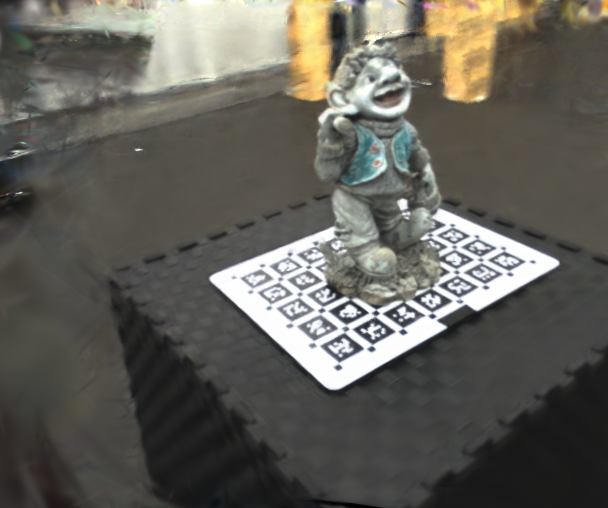}
    \includegraphics[trim=9cm 12cm 8cm 3cm,clip,width=0.49\linewidth]{figs/gaussian_splat/blurry_image.png}
    \includegraphics[trim=11cm 14.05cm 7cm 2cm,clip,width=0.49\linewidth]{figs/gaussian_splat/full_500ms.png}
    \caption{The geometric consistency of the compressed images is demonstrated via the improved 3D Gaussian Splatting reconstruction.}
    \label{fig:gaussian_splat}
\end{subfigure}\\[0.5ex]
\begin{subfigure}{\figmethodwidth}
    \centering
    \includegraphics[trim=0cm 2cm 0cm 4cm,clip,width=0.49\linewidth]{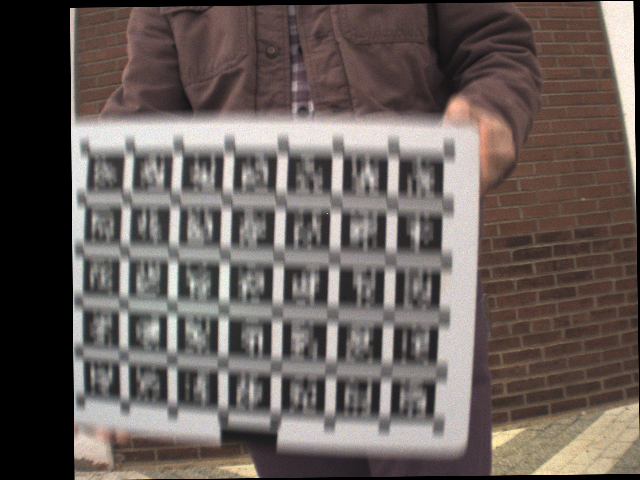}
    \includegraphics[trim=0cm 2cm 0cm 4cm,clip,width=0.49\linewidth]{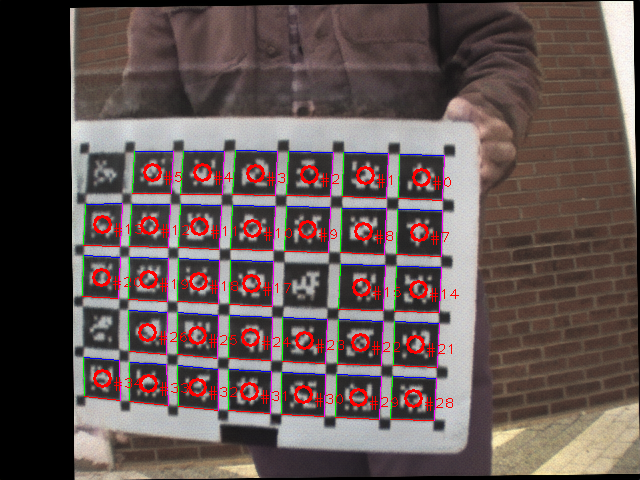}
    \includegraphics[trim=9cm 9.4cm 9cm 5cm,clip,width=0.49\linewidth]{figs/detection/GT.png}
    \includegraphics[trim=9cm 9.4cm 9cm 5cm,clip,width=0.49\linewidth]{figs/detection/kplanes_250ms_tags.png}
    \caption{AprilTag detection at the same moment in time. 
    Detections increase from 0 to 32 by decompressing images using events.}
    \label{fig:tag_correction}
\end{subfigure}
    \caption{\label{fig:enter-label2}Qualitative downstream applications comparing original blurry frames (Left in each subfigure) to the decompressed frames (Right in each subfigure). 
    }
\vspace{-4ex}
\end{figure}

\vspace{-1ex}
\subsection{Downstream Applications}
\label{sec:downstream}
\textbf{Blur-free Gaussian Splatting}.
3D reconstruction has long been a task that has achieved varying results with traditional images and events~\cite{wang2022evac3d}. 
It is also a task commonly plagued by non-ideal input images (e.g. blurry images). 
Here, camera pose estimation and calibration are achieved by running COLMAP \cite{schoenberger2016mvs,schoenberger2016sfm} for the original images and our decompressed images separately. These results are used to reconstruct the 3D scene with 3D Gaussian Splatting~\cite{kerbl3Dgaussians}. Figure~\ref{fig:gaussian_splat} shows the final results with \ours{} providing significantly sharper reconstructions.

\textbf{Fiducial Tag Detection}. 
AprilTags~\cite{olson2011apriltag} represent an external validation of the geometric consistency of our reconstruction at a fine-grained resolution. Each tag is detected through the high-contrast square and decoded through the bit pattern on the interior. This AprilTag grid is commonly used in calibration pipelines~\cite{maye2013self} or as robust SLAM features~\cite{pfrommer2019tagslam}. 
For these tasks, many detections are required. 
We see an improvement in detection rate from 8,854 detections in the original imager to 10,342 with \ours{} (as can be noticed in \cref{fig:tag_correction}).
\Cref{tab:tag} shows the number of detected tags for different methods.
On average, our detection rate increased $20\%$ over interpolation methods.
\begin{table}[tb]
\centering
\adjustbox{max width=.9\linewidth}{
\setlength{\tabcolsep}{6pt}
\begin{tabular}{lcccc}
\toprule
  & Original & FILM \cite{reda2022film} & TimeLens \cite{Tulyakov21cvpr} & Ours\\
  \midrule
  \# tags & 8,854  &  8,853 & 8,628 & \textbf{10,342} \\
\bottomrule
\end{tabular}
}
\vspace{-1ex}
\caption{\label{tab:tag}Total number of fiducial tag detections. 
\ours{} (ours) outperforms the two other VFI baselines and the original image data due to the better sharpness of the reconstructed images.}
\vspace{-1ex}
\end{table}

\vspace{-1ex}
\section{Ablation Studies}
We provide ablation studies on the BS-ERGB dataset to provide insight into the performance impacts of the motion and synthesis modules. 
The branches are complementary to each other. 
Using only the trajectory field warping~\cref{sec:event_flow}, the network can warp sharp pixels from the initial frame, which works better in visible (no occlusion) areas. 
The synthesis module~\cref{sec:synth} provides texture guidance on occluded areas based on events.
We observe in \cref{tab:abl} that the full model outperforms either branch, indicating complementary relationship between the two branches.
\begin{table}[t]
    \centering
    \adjustbox{max width=.9\linewidth}{
    \setlength{\tabcolsep}{6pt}
    \begin{tabular}{lccccc}
    \toprule
    & & \multicolumn{2}{c}{skip 1}& \multicolumn{2}{c}{skip 3}\\
    \cmidrule(l{1mm}r{1mm}){3-4}
    \cmidrule(l{1mm}r{1mm}){5-6}
    &  & PSNR$\uparrow$ & LPIPS$\downarrow$ & PSNR$\uparrow$ & LPIPS$\downarrow$ \\
    \midrule
    Synthesis-only       && 22.90  & 0.191 & 22.36 & 0.219 \\
    Motion-only      & & 23.37  & 0.210 & 22.22 & 0.212 \\
    Full model      & &\textbf{25.40}  & \textbf{0.088} &  \textbf{24.80} & \textbf{0.095}  \\ 
    \bottomrule
    \end{tabular}
    }
    \vspace{-1ex}
    \caption{\label{tab:abl}Ablation studies. 
    \emph{Motion} uses only the motion prediction branch, and \emph{Synthesis} uses only the synthesis branch.}
    \vspace{-3ex}
\end{table}

\vspace{-1ex}
\section{Conclusion}
We have presented \ours{}, a novel method for event-based continuous color video decompression, using a single static RGB image and following events. 
The core of the method is founded on two novel representations for both the photometric changes and the pixels' motions predicted from events. 
Our approach does not rely on additional frames except for the initial image, enhancing robustness to sudden lighting changes, minimizing prediction latency, and reducing bandwidth requirements. 
We thoroughly evaluate our method on a standard dataset and our more challenging dataset, \ourdata{}, showing state-of-the-art performance in event-based video color video decompression. 
Additionally, we showcased practical applications of our method in various scenarios, including 3D reconstruction and camera fiducial tag detection, even under challenging lighting and motion conditions.

\fi %

\ifarxiv
  \textbf{Acknowledgment:} We gratefully acknowledge the support by the following grants: NSF FRR 2220868, NSF IIS-RI 2212433, NSF TRIPODS 1934960, ONR N00014-22-1-2677, the Deutsche Forschungsgemeinschaft (DFG, German Research Foundation) under Germany’s Excellence Strategy – EXC 2002/1 ``Science of Intelligence'' – project number 390523135 and the NSF AccelNet NeuroPAC Fellowship Program. We thank Jiayi Tong for help with visualization and proofreading.

  \ifshowsupplementary
  \section*{SUPPLEMENTARY MATERIAL}
  
\Cref{sec:supp:imp_details} provides implementational details of our and the baseline methods.
\Cref{sec:supp:dataset} presents more details on \ourdata{} and our camera setup.

\section{Implementation Details}
\label{sec:supp:imp_details}
In this section, we provide full details of the proposed architecture (\cref{sec:supp:model}), specifics of the training (\cref{sec:supp:training}), a discussion of the input representation (\cref{sec:supp:input_rep}), and lastly details on the implementation of the baseline methods (\cref{sec:supp:details_baselines}).

\subsection{Model}
\label{sec:supp:model}

\paragraph{K-plane Synthesis Network.}
The K-plane synthesis network has three components: feature encoder, K-plane sampler, and pixel color decoder.
The feature decoder is implemented as a U-Net with 4 encoders, 2 residual bottleneck layers, and 4 decoders, with skip connections. The U-Net backbone is depicted in Figure~\ref{fig:backbone}. The feature encoder maps the input event volumes and an initial image to a $24$ channels, $8$ for each plane. The K-plane sampler uses bilinear sampling to query features at a given $x, y, t$ coordinate. The queried features are concatenated and fed into a final pixel color decoder. The pixel color decoder can be implemented using either a lightweight MLP or a lightweight CNN. For our task, we often care about a whole image rather than individual pixels. Therefore, we deployed an efficient three-layer CNN decoder, with each hidden layer having 64 kernels and ReLU activation, to decode image pixel color.

\paragraph{Motion Basis Network.}
The motion basis network is a MLP network that maps a single scalar time value to a set of query values. This network is shared between \textbf{all} instances, which means that it is not conditioned on the input of events and images into the network. We use it as a set of query-able, time-continuous functions. The network has two hidden layers with 64 neurons and ReLU activation functions. The network outputs 5 values which are the output of 5 basis functions. The function values are also shared between $x$ and $y$ trajectories. The $x$ and $y$ trajectory coefficients are predicted separately per-pixel, as detailed in the next section.

\begin{figure}
    \centering
    \includegraphics[trim=3.5cm 5.7cm 0.5cm 5.0cm,clip,width=1.5\linewidth]{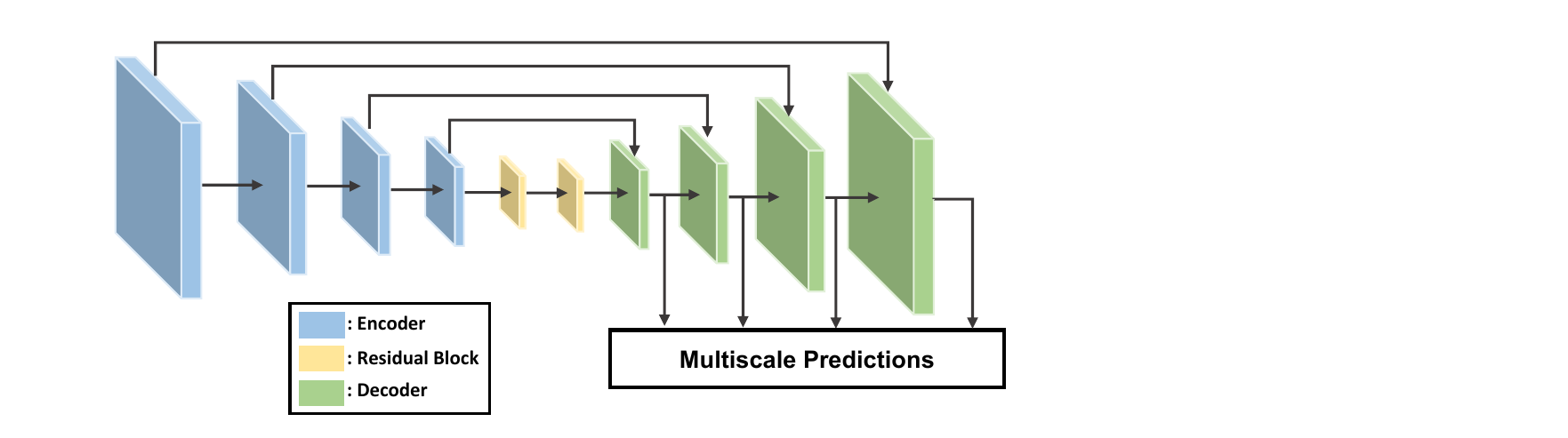}
    \caption{Backbone architecture. The U-Net with skip connections maps event volumes and images into multi-scale dense output.}
    \label{fig:backbone}
\end{figure}

\begin{figure*}
    \centering
    \includegraphics[trim=1.5cm 7.1cm 0.5cm 5.0cm,clip,width=1\linewidth]{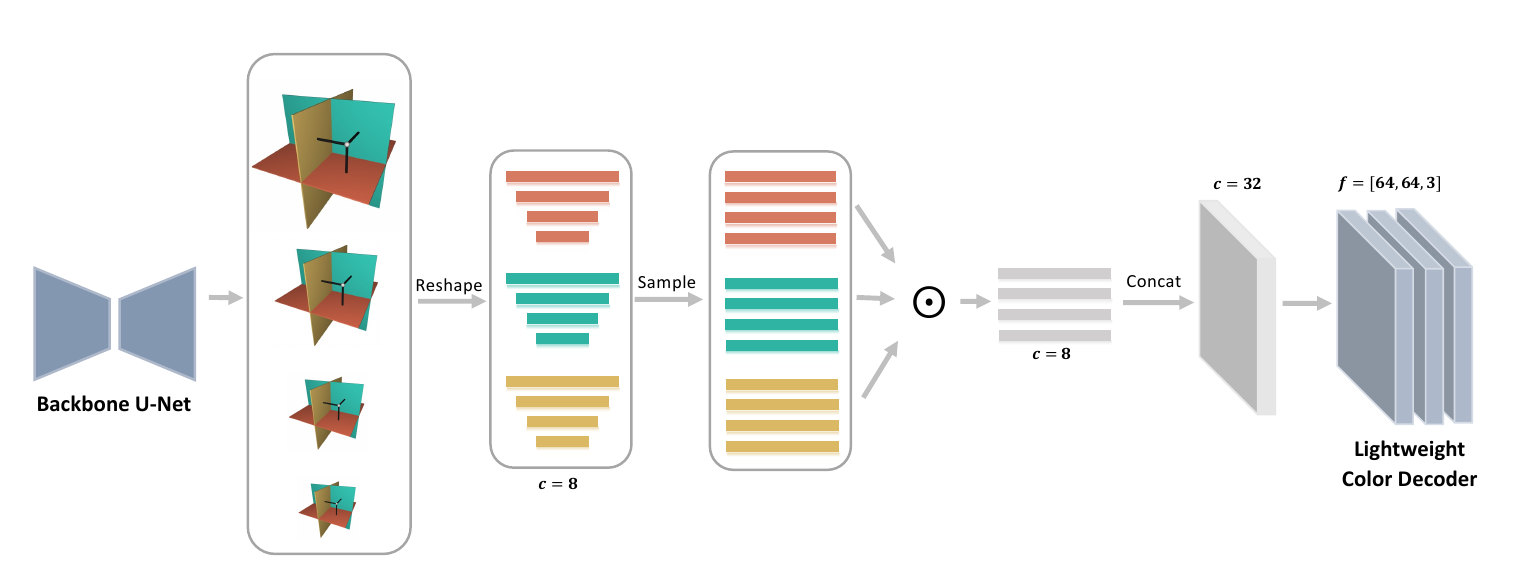}
    \caption{The architecture for the K-planes synthesis module. The initial image and event volume are mapped to multi-scale Tri-planes (xy, xt, yt). These feature planes are sampled bilinearly and fed into a lightweight color decoder network.}
    \label{fig:supp_kplanes}
\end{figure*}

\paragraph{Motion Trajectory Field Network.}
The motion trajectory network is based on the same backbone as the K-plane synthesis branch, as described in \cref{fig:backbone}. The network predicts $(K \times 2 + 1)$ channels. The first $2K$ channels are the motion coefficients for x and y trajectory separately. The last channel is the splatting weight for the Softmax splatting operation. We use Tanh activation for the Softmax weight to improve the numerical stability of the Softmax function.

\paragraph{Multi-scale Feature Fusion Networks.}
We include the network architecture of our multiscale feature fusion network in \cref{fig:fusion}. The warped features at different scales are upsampled to the nearest-neighbor method. The features are then fed through a series of convolutional layers. By doing this iteratively, we gradually fuse the multi-scale image and feature pyramids in a course-to-fine fashion.

\begin{figure}
    \centering
    \includegraphics[trim=1.5cm 6cm 18.5cm 4.0cm,clip,width=0.8\linewidth]{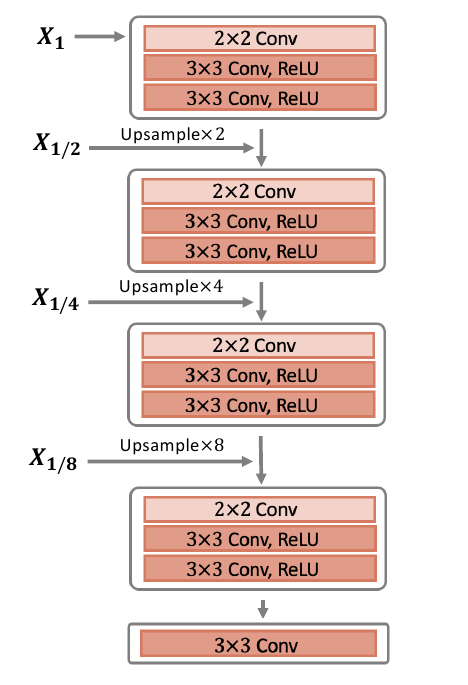}
    \caption{Architecture for the multi-scale feature fusion network. The warped feature and image pyramids are gradually injected into the network via a series of upsampling and convolution.}
    \label{fig:fusion}
\end{figure}

\subsection{Training}
\label{sec:supp:training}

We train the K-plane synthesis network and the motion network separately and jointly optimize a shared feature fusion network to fuse the two branches.

For the K-plane synthesis network, we use a learning rate of $10^{-4}$ with an Adam optimizer. We train the network for 20,000 iterations. We separately train the motion network to predict pixel trajectories, supervised with a image warping loss and L1 flow loss described in Section~\cref{sec:event_flow}.
Additionally, we train a dummy feature fusion network to map the warped image features and dummy synthesized frames to a final prediction. Here, we do not have the latent frame flow for warping features. The key idea is to train the network to rely only on events to estimate motion trajectories. The network is trained for 15,000 iterations with a learning rate of $10^{-4}$ with an Adam optimizer. We directly use FILM's~\cite{reda2022film} pre-trained multi-level feature encoder without fine-tuning.

In the end, we take the trained K-plane synthesis network and motion network and fuse their predictions with a multilevel feature fusion network. In this final training, we insert the warped features according to the latent frame flow as described in~\cref{sec:latent_flow}. 
We jointly optimize all parts of the network except the frozen image encoder. The entire network is trained 100,000 steps with the same learning rate and optimizer configuration as the motion network. We design the training process to maximize the information learned in the K-plane synthesis network and the motion network. For data augmentation, we randomly perform flipping along the x and y directions, randomly rotate at $[0, 90, 180, 270]$ degrees, and then randomly rotate between $(-45, 45)$ degrees.

\subsection{Input Representation}
\label{sec:supp:input_rep}
We use an event volume similar to EV-FlowNet~\cite{Zhu18rss}. For a set of events $\mathbf{\Gamma} = \{(x_i, y_i, t_i, p_i)\}$, an event volume $E(x,y,t)$ containing these events is written as
\begin{align}
E(x, y, t) = \sum_i p_i k_b(x-x_i)k_b(y-y_i)k_b(t-t_i),
\end{align}
where $k_b(\cdot)$ is a bilinear sampling kernel function. We used 60 temporal channels to encode events into spatio-temporal volumes. A large number of input channels is selected because we encode the long-range event motion. For the event-based motion network, we normalized the event volume to have a maximum of 1 and a minimum of 1 because the network flow should be invariant to the absolute number of events in the volume. For the synthesis branch, we do not normalize because the pixel-wise change should vary with respect to the number of events. We separated positive and negative events and concatenate the volumes together because of the asymmetric contrast threshold. Positive and negative events cannot be canceled out by each other due to the unknown ratio between the two contrast thresholds.

\subsection{Baselines}
\label{sec:supp:details_baselines}

This section provides additional information to the baseline methods.

\textbf{Frame Interpolation.} We use FILM ~\cite{reda2022film} for direct frame interpolation between two key frames.
This method does not use event data and relies on the key frame before and after the desired generated frame.

\textbf{Video Prediction.} DMVFN ~\cite{hu2023cvpr} performs video prediction.
The model takes two consecutive frames and predicts the next one.
We predict the hold-out frames after a key frame, in an iterative manner, starting with a key frame and the original frame before the key frame.

\textbf{Flow Unrolling + RAFT.} In this method, the frame is iteratively warped forward in multiple steps using optical flow, which is determined using RAFT~\cite{Teed20eccv}.
To allow the warping frame $f_{i}$ to $f_{i+1}$, we determine the backward flow from $t_{i+1}$ to $t_{i}$, using the original frames (ground truth).

\textbf{Flow Unrolling + E-RAFT.}
This method is similar to the previous method.
However, the flow is obtained from E-RAFT~\cite{Gehrig21threedv}, which outputs the optical flow between two timestamps $t_1$ and $t_2$, by using events in a window before $t_1$ and the events between the two timestamps.
This method uses a frame and the following events and is therefore closest to our method with respect to the data it uses.

\textbf{Inpainted Flow unrolling.} For both flow unrolling methods, we additionally refine the predictions by training a U-Net to reconstruct the original frames from the intermediate (flow unrolling) predictions.
We use different models for the results on BS-ERGB and \ourdata{}, trained on the respective training splits with a learning rate $\lambda = 0.001$, a batch size of 8 for 100 epochs.

\section{Dataset Details}
\label{sec:supp:dataset}

This section presents the single objective beam splitter design and details on \ourdata{}.

\subsection{Beam Splitter Design}

With the lack of integrated APS pixels (that was a common feature in prior cameras~\cite{Brandli14ssc,Taverni18tcsii}) within recent event based cameras, beam splitters have become common practice to achieve zero baseline images and events~\cite{Shiba23pami,Wang20cvpr, Hidalgo22cvpr,Tulyakov21cvpr, wang2024continuous}. High resolution imagers paired with event based cameras provide the best case scenario for sensor fusion allowing high quality labels such as pixel intensity or semantics. Previous beam splitters constructed for event based camera systems leveraged multiple objective lenses and require a full intrinsic and extrinsic calibration for warping between the imager and event based camera. Mounting a beam splitter after the objective lens allows the sensors to share the same distortion and projection function. We migrate away from c-mount to f-mount lenses in order to achieve the flange distance required to fit a beam splitter cube into the optical path. \Cref{fig:beam-splitter} shows the physical layout of the beam splitter system. The distance between the flange and sensors are the critical distance: comprised of the distance between the sensors and the beam splitter as well as the distance from the beam splitter to the flange. Our system was 3D printed and adjusted for the printer to achieve the ideal back focus.

\begin{figure}[t]
    \centering
    \includegraphics[width=0.6\linewidth]{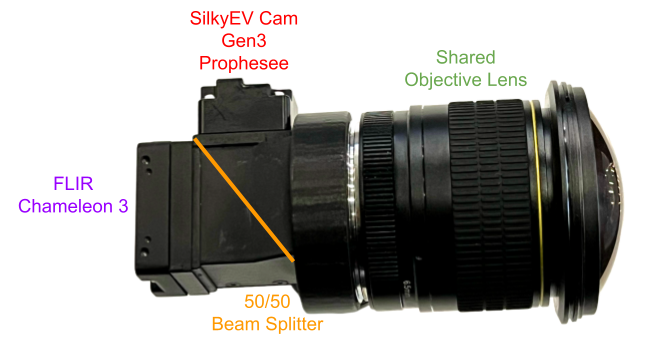}
    \caption{Constructed single objective beam splitter with each major component labeled. As constructed the Event Camera will be flipped compared to a traditional setup.}
    \label{fig:beam-splitter}
\end{figure}

\subsection{Dataset Design}

\begin{table}[t]
    \centering
    \begin{tabular}{c|c|c}
         Brand & Model & Qty \\ \hline
         ThorLabs & CCM1-BS013 & 1 \\
         CenturyArks & SilkyEvVGA & 1 \\
         FLIR & Chameleon 3 & 1 \\
         Opteka & 6mm F-Mount Lens & 1 \\
          & C to F Mount Adapter  & 1 \\
          & M2 Threaded Insert & 12 \\
    \end{tabular}
    \caption{Beam splitter bill of materials}
    \label{tab:bill_of_materials}
\end{table}

We collect data from the beam splitter with CoCapture~\cite{Hamann22icprvaib,Shiba23pami}. The raw data was temporally aligned using a trigger signal provided by a micro-controller.
We calculate the frame timestamps as the time of the trigger events plus half of the exposure time.
We follow~\cite{Muglikar21cvprw, Chaney23cvprw, wang2022evac3d} for the camera calibration.
A set of matching corner points is extracted from pairs of images and event intensity images (from e2vid~\cite{Rebecq19pami}).
The set of matching corner points is used to directly determine the homography matrix by minimizing the reprojection error between the two domains.
Using the calibration, the frames are warped into the event domain.

Our dataset is divided into training, validation, and testing by recording.
It is curated such that all frames in the three main parts (training, validation, testing) have sharp frames without motion blur.
This is important for training as well as quantification of reconstruction errors.
However, event-based cameras are able to operate in regimes that frame-based cameras cannot.
To this end, we provide an additional set of sequences for qualitative comparison only.
These recordings contain challenging scenarios, where only a small subset of the frames is sharp and the rest underlies heavy motion blur.
The subset contains recordings for the downstream tasks novel-view synthesis and tag detection that were shown in~\cref{sec:downstream},
and further examples for human-pose estimation and rapid camera motion.
These additional performance categories provide information on where methods excel and fail.

We will release all full raw sequences in addition to the calibrated and aligned data for future work at the highest quality and with the greatest flexibility. The structure of the raw data will not include a data split as the split we have chosen is optimal for reconstruction purposes, but not necessarily for all tasks.

  \fi
  {
    \small
    \bibliographystyle{ieeenat_fullname}
    \bibliography{all,main}
  }

\else

  \textbf{Acknowledgment:} We gratefully acknowledge the support by the following grants: NSF FRR 2220868, NSF IIS-RI 2212433, NSF TRIPODS 1934960, ONR N00014-22-1-2677, the Deutsche Forschungsgemeinschaft (DFG, German Research Foundation) under Germany’s Excellence Strategy – EXC 2002/1 ``Science of Intelligence'' – project number 390523135 and the NSF AccelNet NeuroPAC Fellowship Program. We thank Jiayi Tong for help with visualization and proofreading.

  {
    \small
    \bibliographystyle{ieeenat_fullname}
    \bibliography{all,main}
  }

  \ifshowsupplementary
  \clearpage
  \maketitlesupplementary
  
  \fi

\fi %

\end{document}